\documentclass{styles/svproc}
\usepackage{longtable} % Damit können Tabellen über einen Seitenwechsel umgebrochen werden.
\usepackage{multirow} % ermöglicht mehrere Zeilen in einem Tabellenfeld zu nutzen
\usepackage{framed} 
\usepackage{booktabs, colortbl}
\usepackage{url}
\usepackage{graphicx}

\usepackage{siunitx}
\usepackage{caption}
\usepackage{subcaption}
\usepackage{wrapfig}
\usepackage{cleveref}

\usepackage{todonotes}
\usepackage{soul} % text marking

\newcommand{\ra}[1]{\renewcommand{\arraystretch}{#1}}

\makeatletter
 \if@todonotes@disabled
 
 \else
 
 \fi
\makeatother

\begin{document}
\mainmatter              % start of a contribution
\title{Connected Dependability Cage Approach for Safe Automated Driving}
\titlerunning{Connected Dependability Cage Approach for Safe Automated Driving}  % abbreviated title (for running head)
\author{Adina Aniculaesei \and Iqra Aslam \and Daniel Bamal \and Felix Helsch \and Andreas Vorwald \and Meng Zhang \and Andreas Rausch}
\authorrunning{A. Aniculaesei et al.} % abbreviated author list (for running head)
\tocauthor{Adina Aniculaesei, Iqra Aslam, Daniel Bamal, Felix Helsch, Andreas Vorwald, Meng Zhang, and Andreas Rausch}
\institute{Institute for Software and Systems Engineering, TU Clausthal,\\ Clausthal-Zellerfeld, 38678, Germany \\ 
\email{\{adina.aniculaesei, iqra.aslam, daniel.bamal, felix.helsch, meng.zhang, andreas.rausch\}@tu-clausthal.de}, \email{andreas.vorwald@gmail.com}}
%\\CARIAD Technology, Wolfsburg, 38442, Germany}

\maketitle              % typeset the title of the contribution
\begin{abstract}

Automated driving systems can be helpful in a wide range of societal challenges, e.g., mobility-on-demand and transportation logistics for last-mile delivery, by aiding the vehicle driver or taking over the responsibility for the dynamic driving task partially or completely. Ensuring the safety of automated driving systems is no trivial task, even more so for those systems of SAE Level 3 or above. To achieve this, mechanisms are needed that can continuously monitor the system's operating conditions, also denoted as the system's operational design domain. This paper presents a safety concept for automated driving systems which uses a combination of onboard runtime monitoring via connected dependability cage and off-board runtime monitoring via a remote command control center, to continuously monitor the system's ODD. On one side, the connected dependability cage fulfills a double functionality: (1) to monitor continuously the operational design domain of the automated driving system, and (2) to transfer the responsibility in a smooth and safe manner between the automated driving system and the off-board remote safety driver, who is present in the remote command control center. On the other side, the remote command control center enables the remote safety driver the monitoring and takeover of the vehicle's control. We evaluate our safety concept for automated driving systems in a lab environment and on a test field track and report on results and lessons learned.
% We would like to encourage you to list your keywords within
% the abstract section using the \keywords{...} command.

\end{abstract}
%

% ------------------------ BEGIN SECTION -----------------------------
\section{Introduction}
\label{sec:introduction}
Automated driving systems (ADSs) have become more present in a variety of applications that address current societal challenges, e.g., mobility-on-demand and last-mile delivery logistics, by assisting the driver to carry out the dynamic driving task (DDT) or taking over the responsibility for the DDT partially or completely. Ensuring that automated driving systems operate safely both for the system and its environment is not a trivial task. 

The standard SAE J3016 \cite{sae3016.2021} defines six levels of automation for automotive systems, from SAE Level 0 (SAE L0) to SAE Level 5 (SAE L5). The first three levels of automation refer to driver support features, with the driver being in charge of supervising and partially carrying out the DDT as well as supervising the vehicle’s environment. For ADSs of SAE L1 and L2, it is important that in case the system reacts, then its reaction must be correct. The safety requirements of the ADS are the main focus. The system behavior is designed to be conservative in order to build the system to be fail-safe.  

Starting with SAE L3, the ADS is in charge of executing the DDT and of supervising the vehicle’s environment. Automated driving systems with SAE L3 and L4 are activated and can execute the DDT only when certain operating conditions are satisfied. In case the operating conditions are not satisfied anymore, the driving system requires the intervention of the driver. While at SAE L3, the driver is still required to be ready to intervene and take over control of the vehicle, starting with SAE L4 the ADS must be ready to trigger the necessary measures that can bring the vehicle to a safe state, e.g., pulling on the side of the road. For systems of SAE L3 and above, it is important that the system reacts in all situations and its reaction must be correct. In this case both safety and liveness requirements are in focus and the goal is to make the system fail-operational.

Various methods for verification and validation are needed in order to ensure that ADSs of SAE L3 or above can operate safely in a realistic road environment. Automated driving systems undergo extensive assessment to demonstrate compliance with functional safety (FuSa) standards, such as ISO 26262 \cite{iso26262.2011}. However, conventional safety standards are no longer sufficient for the next generation of ADSs and for fully automated driving. Complementary to ISO 26262, the standard ISO 21448 \cite{iso21448.2022} aims for the safety of the intended functionality (SOTIF) of the ADS, which is equivalent to the absence of unreasonable risk due to hazards resulting from functional insufficiencies. These insufficiencies result from the ADS operating in an environment which dos not comply with its operational design domain (ODD) specification. Thus, in addition to methods that ensure compliance with FuSa, innovative approaches are needed to demonstrate SOTIF for ADSs of SAE L3 and above. One approach that contributes to ensuring and demonstrating SOTIF is runtime monitoring of the ODD. 

This paper proposes an integrated safety concept for ADSs centered around the notion of connected dependability cage, which is able to monitor the safety requirments of the ADS during the system operation in it environment. This safety concept extends the concept of dependability cage, first introduced in \cite{Aniculaesei.2018} and then refined in \cite{Grieser.2020}. In its initial concept, a dependability cage consists of two main components: a qualitative monitor and a quantitative monitor (cf. \cite{Aniculaesei.2018}, \cite{Grieser.2020}). The qualitative monitor checks during the system operation the correctness of the system behavior with respect to the defined safety requirement specification (cf. \cite{Aniculaesei.2018}). If the qualitative monitor detects a violation of the safety requirement specification, then this result is recorded in a knowledge database (cf. \cite{Grieser.2020}). In turn, the quantitative monitor evaluates during system operation the current driving situation of the ADS and checks whether the system is still in a context that was verified through various methods, e.g., system testing, during the system's design time (cf. \cite{Aniculaesei.2018}). If it has not been tested at design time, then the current driving situation is logged in a knowledge database as a novelty situation that occurred during system operation (cf. \cite{Aniculaesei.2018}). The results of the qualitative monitor and the quantitative monitor are used in a two-folded manner. In case of warnings from the two monitors, these results are used to compute possible reactions of the system that can bring the system back in a safe state, e.g., emergency braking. On the other side, these results are used in further development iterations to improve the system development artifacts during system design, e.g., better test cases to improve the test coverage for testing the qualitative monitor or better training data for the training of the quantitative monitor. 

The safety concept proposed in this paper consists of a connected dependability cage and a remote command control center (remote CCC). The runtime monitoring of a system's ODD occurs onboard the ego-vehicle and off-board. The connected dependability cage monitors the ODD onboard the ego-vehicle using input data from its sensors. The off-board monitoring is done by a remote safety driver which supervises the ego-vehicle through the remote CCC. This safety concept is realized through a modular software architecture which allows reconfiguration of the ADS based on the monitoring results of the connected dependability cage and the instructions given by the remote safety driver from the remote CCC. The safety concept is evaluated in a lab environment using a model car and on a test field track with a full-size vehicle. The use case scenario used for the concept evaluation pertains to the application domain of parcel delivery logistics and was defined together with academic and industry partners in the project VanAssist. 

The rest of the paper is structured as follows. Section \ref{sec:related-work} researches relevant related work. In Section \ref{sec:solution-concept}, the integrated safety concept for ADSs is presented in detail. Section \ref{sec:case-study} introduces the case study and the project VanAssist. The evaluation in the lab environment and on the test field track is presented in detail in Section \ref{sec:evaluation}. Section \ref{sec:summary} concludes this paper and points out to interesting future research directions.

% ------------------------ END SECTION -------------------------------

% ------------------------ BEGIN SECTION -----------------------------
\section{Related Work}
\label{sec:related-work}
Our brief literature research is focused around methods for runtime monitoring of properties for autonomous safety-critical systems, safety architectures for safety-critical applications and approaches that use the concept of safety cage to ensure the system safety. 

Schirmer et al. \cite{Schirmer.2023} discuss the challenges of monitoring safety properties of autonomous aircraft systems, including those that involve temporal and spatial aspects. The authors recognize the need for runtime safety monitors to be integrated with the system under analysis and thus to have access to the overall system. Furthermore, they propose that the monitoring properties follow the hierarchy of the system under analysis. Thus, different monitoring properties can be formulated at different system hierarchy levels. They focus on the hierarchy levels introduced in the SAE standard ARP4761, i.e., item, system, and aircraft (cf. \cite{arp4761.1996}), and extend these to include mission and operation levels for autonomy. The monitoring properties are classified in different categories, i.e., temporal, statistical, spatial and parameterized, and different formal specification languages are used to formalize properties situated at different levels in the system hierarchy (cf. \cite{Schirmer.2023}). 

The integration of the runtime safety monitors with the system under analysis must be supported by the system safety architecture. The access of the runtime monitors to the overall system can be ensured only through appropriate interfaces between the monitors and the system under analysis. Various safety architectures have been proposed over the years for automated safety-critical systems. A well-known safety architecture is the Simplex architecture, introduced by Sha in \cite{Sha.2001}. The system has a high-assurance controller and a high-performance controller, which can fulfill the task of the system independent of each other, as well as a decision module that monitors the system state. The decision state switches from the high-performance controller to the safety controller whenever the system approaches an unsafe state (cf. \cite{Sha.2001}). 

Jackson et al. \cite{Jackson.2021} introduces Certified Control, a variation of the Simplex architecture. A monitor checks the actions of the main controller before forwarding them to the actuators and blocks any action that is considered unsafe or replaces it with a safer action (cf. \cite{Jackson.2021}). The decision to block an action of the main controller is taken based on a certificate generated by the latter. This certificate contains evidence that the proposed action is safe. Once the certificate is approved by the monitor, the action of the man controller is forwarded to the actuators (cf. \cite{Jackson.2021}). The concept of Certified Control is illustrated with a certificate for LiDAR data and its formal verification through a Hoare-style proof carried out by hand (cf. \cite{Jackson.2021}). In \cite{Bansal.2022}, Bansal et al. propose Synergistic Redundancy as a safety architecture for complex cyber-physical systems (CPS), e.g., autonomous vehicles (AV). The Synergistic Redundancy architecture decouples the mission layer from the safety assurance layer of the system. The mission layer executes all tasks necessary to fulfill the system mission, e.g., perception, planing, and control. The safety layer runs in parallel to the mission layer and communicates over predefined interfaces with the mission layer. The safety layer  provides algorithms for deterministic guarantees as well as fault handlers that identify faults and take corrective actions (cf. \cite{Bansal.2022}). The Synergistic Redundancy concept is demonstrated for the safety-critical function of obstacle detection and collision avoidance (cf. \cite{Bansal.2022}). Phan et al. \cite{Phan.2017} present a component-based variant of the Simplex architecture, to ensure the runtime safety of component-based CPSs. The proposed approach combines the principles of the Simplex architecture with assume-guarantee reasoning in order to formally prove system guarantees with respect to energy safety, collision freedom and mission completion for a ground rover (cf. \cite{Phan.2017}). 

Considerations about the safety architecture of an automated safety-critical system become even more important when part of the system functionality is realized with artificial intelligence (AI) or machine learning (ML) components. Fenn et al. \cite{Fenn.2023} take a closer look at common architectural patterns used in traditional aviation systems and discuss the implications for the safety assurance of the whole system when AI/ML components are integrated in the system architecture. 

In \cite{Costello.2023}, Costello and Xu propose a new approach to certifying the safety of autonomous systems in the naval aviation domain. The proposed safety architecture consists of a runtime assurance (RTA) input monitor and a controller/safety monitor. The current aircraft state and a projection of the aircraft state into the future are passed as inputs to the RTA input monitor, which processes these further for the safety monitor. In turn, the safety monitor determines if the aircraft will violate the clearance envelope for autonomous behavior. If the aircraft violates the clearance envelope, then the safety monitor switches the air vehicle guidance to deterministic behavior.  

Borg et al. \cite{Borg.2023} use a safety cage to carry out validity and safety checks for an ML-based pedestrian automatic emergency braking system, called SMIRK, whose task is to detect pedestrians and avoid any collisions with them. The safety cage receives radar/LiDAR and camera input data and produces an assessment whether a collision with a pedestrian is imminent or not. On one side, the safety cage uses an ML-trained anomaly detector to analyse the input camera images with potential pedestrians detected in order to find any anomalies with respect to its training data (cf. \cite{Borg.2023}). On the other side, the safety cage performs uses a rule engine to do heuristics-based sanity checks, e.g., in order to determine if the perceived situation is consistent with the laws of physics (cf. \cite{Borg.2023}). The authors use  SMIRK as an example system to demonstrate the systematic construction of a safety case, including the system architecture, the safety requirements, and the test scenarios used to ensure the safety of the system (cf. \cite{Borg.2023}). 

Our paper builds on a foundation of research developed in several previous publications. The concept of dependability cage was first proposed in \cite{Aniculaesei.2018} together with the challenges of engineering hybrid AI-based ADSs that emerge with respect to the dependability and safety assurance of these systems. This concept has already been applied on a lane change assistance system (LCAS) (cf. \cite{Mauritz.2014}, \cite{Mauritz.2015}, \cite{Mauritz.2016}, \cite{Mauritz.2020}).  

Recently, the concept of the connected dependability cage has been introduced as an extension of the initial notion of dependability cage (cf. \cite{Helsch.2022}). Its application in a scenario of parcel delivery logistics in the project VanAssist has been described in our previous work in \cite{Helsch.2022}. In \cite{Helsch.2022}, the focus was placed on improving the algorithm for the computation of the safe zone around the ego-vehicle, in comparison to the one used in \cite{Grieser.2020}, in order to address the challenges in the project VanAssist. For additional details on how the safe zone of an AV is defined, the reader is referred to Section \ref{sec:solution-concept} of this paper. %\todohl{Safe Zone}{If the Safe Zone is note yet introduced, we could add a Footnote here to introduce it, or refer to a later section}.

Compared to our previous work in \cite{Helsch.2022}, in this paper we describe in more detail the purpose and functionality of the remote CCC, also developed in the VanAssist project, as well as the mechanism which enables the seamless share and transfer of responsibility over the DDT between the ADS and the remote safety driver in the remote CCC.

% ------------------------ END SECTION -------------------------------

% ------------------------ BEGIN SECTION -----------------------------
\section{Safety Concept for Automated Driving Systems via Connected Dependability Cage}
\label{sec:solution-concept}
\begin{figure}
	\centering
	\includegraphics[width=\textwidth, keepaspectratio]{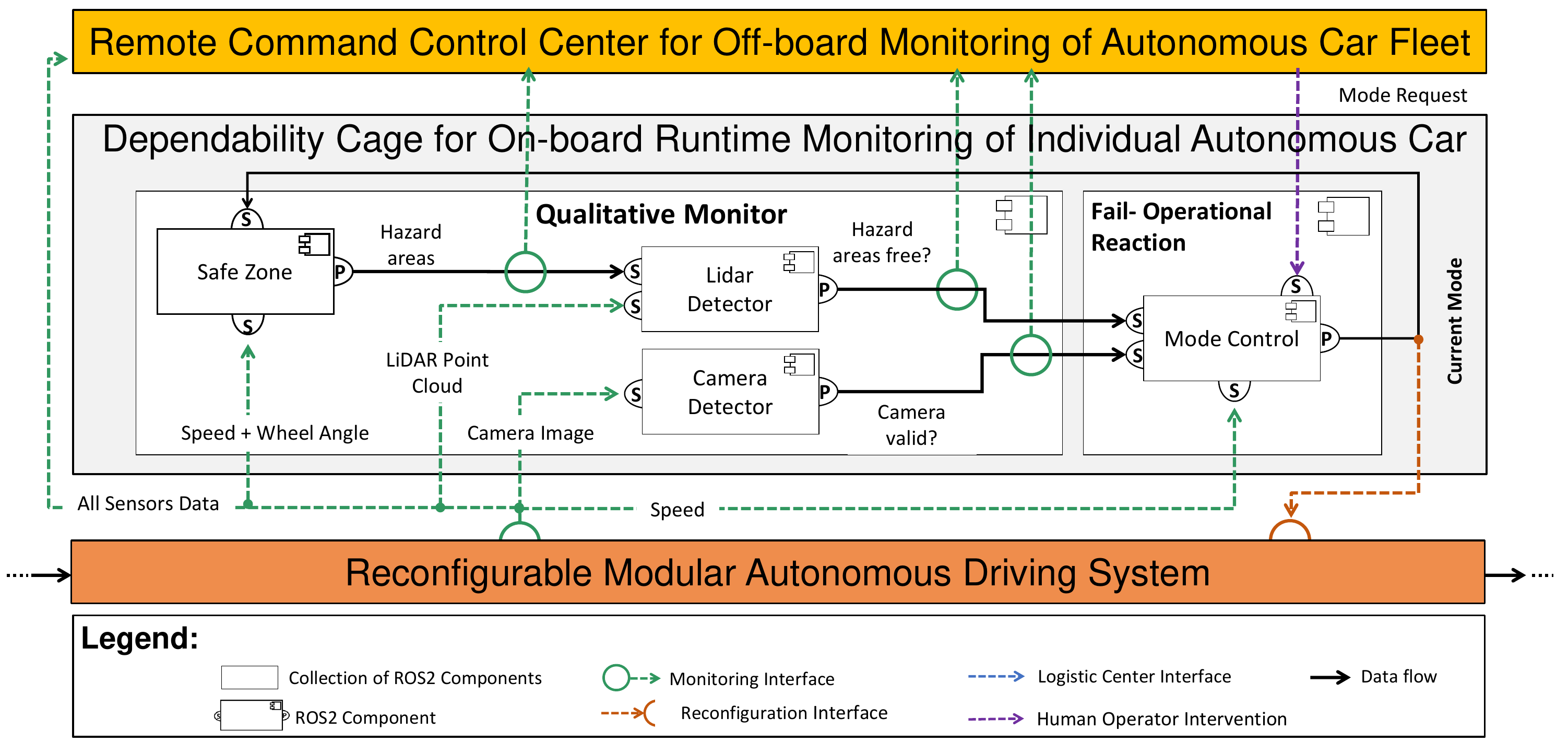}
	\caption{Integrated Safety Concept for ADSs using the Connected Dependability Cage Approach.}
	\label{fig:dc-connected}
\end{figure}

The approach of connected dependability cage is depicted in Figure \ref{fig:dc-connected} and brings together two main systems: (1) an onboard runtime monitoring system of the ADS through the connected dependability cage and (2) an off-board runtime monitoring system through the remote CCC and a human remote safety driver. 

\subsection{Onboard Runtime Monitoring of ADSs with the Connected Dependability Cage}

The connected dependability cage has two major components: (1) a qualitative monitor, which detects the violation of the ADS's safety requirements and (2) a mode control component in charge of the fail-operational reaction of the automated driving system in case the qualitative monitor detects a safety requirements violation. Notice that, in comparison to the initial dependability cage concept, the connected dependability cage presented in this paper does not include the quantitative monitor, as this component has not been implemented in the VanAssist project.

\subsubsection{Qualitative Monitor.} There are two safety requirements formulated for the ADS, which the qualitative monitor must continuously check during the system operation:

\textit{\textbf{SR1:} The ADS shall not cause a collision of the ego-vehicle with static obstacles in the vehicle's environment.} 

\textit{\textbf{SR2:} The ADS shall operate only if the image data provided by the ego-vehicle's camera sensor is valid.}

In order to check the safety requirements SR1 and SR2 during system operation, three components are implemented in the qualitative monitor: (1) a component which computes a safe zone around the ego-vehicle, (2) a LiDAR detector, and (3) a camera validator. The LiDAR detector is used to monitor SR1, using as input the computed safe zone and the data provided by the LiDAR sensors of the ego-vehicle. The safe zone is computed based on the current velocity and steering angle of the ego-vehicle. It consists of two separate areas, denoted as clear zone and focus zone, with the focus zone being computed on top of the clear zone as a constant positive overhead, and therefore always larger than the clear zone. These areas mark danger zones around the ego-vehicle based on its braking path. The LiDAR detector monitors SR1 by checking whether there are obstacles in the clear zone or in the focus zone. If the focus zone is free of obstacles, then the clear zone is also free of obstacles. In turn, the camera validator is used to monitor the safety requirement SR2. This component validates the camera sensor data by quantizing the sharpness of a camera image. If its sharpness falls below a given threshold value, the input image is classified as invalid. 

\subsubsection{Mode Control.} The mode control component triggers a fail-operational reaction, in case the qualitative monitor detects the violation of at least one of the two safety requirements formulated in the previous section. To compute the appropriate fail-operational reaction, the mode control component takes as inputs the results of the LiDAR detector and of the camera validator as well as the requests for change of the cage mode and of the driving mode received from the CCC. The computed fail-operational reaction consists of a new cage mode and a new driving mode. The dependability cage has two modes: on and off. In turn, the automated driving system has five driving modes: 

\begin{itemize}
	\item \textit{Fully Autonomous Driving} represents an autonomous driving function without restrictions, but with stricter safety criteria, e.g., wider safe zone around the ego-vehicle. 
	\item \textit{Limited Autonomous Driving} triggers an autonomous driving function that is restricted in its freedom, e.g., driving with reduced velocity, but is safeguarded by weakened safety criteria, e.g., smaller safe zone around the ego-vehicle.
	\item \textit{Remote Manual Driving} represents driving by a human remote safety driver. 
	\item \textit{In-Place Manual Driving} is driving by a safety driver present in the car. 
	\item \textit{Emergency Stop} implements a driving function that triggers emergency braking on the ego-vehicle.
\end{itemize}

The responsibility for dynamic driving task during the operation of the ego-vehicle is shared between the human safety driver and the ADS. Depending on the driving mode computed by the mode control component, the responsibility for the DDT is carried either by the safety driver or by the ADS individually, or the safety driver shares the responsibility for the DDT cooperatively with the ADS. Thus, the ADS is responsible for carrying out the DDT on its own when the driving mode is \textit{Fully Autonomous Driving}. 

The safety driver is in charge of the DDT when the driving mode is set to \textit{Remote Manual Driving}, \textit{In-Place Manual Driving}, or \textit{Emergency Stop}. The driving mode \textit{Emergency Stop} can be requested by the remote safety driver via the control panel of the remote CCC. It can also be triggered when the cage mode is on and the qualitative monitor has detected a violation of at least one of the two system safety requirements. The release of the emergency brake can be performed only by the safety driver, via a request for one of the other four possible driving modes, i.e., \textit{Remote Manual Driving}, \textit{In-Place Manual Driving}, \textit{Limited Autonomous Driving}, or \textit{Fully Autonomous Driving}. 

The safety driver shares the responsibility of the DDT with the ADS when the driving mode is set to \textit{Limited Autonomous Driving}. This is because adjusting the parameters of the autonomous driving system order to restrict its freedom as well as weakening its safety criteria requires the careful oversight of the remote safety driver. The safety driver can request the driving mode \textit{Limited Autonomous Driving} from the control panel of the remote CCC.

The mode control component is designed as a SCADE state machine using the ANSYS SCADE tool chain. This way, we ensure a verifiable safe transfer of responsibility of the DDT and a smooth cooperation between the ADS and the remote safety driver.

\subsection{Off-board Runtime Monitoring of ADSs through the Remote Command Control Center}

%\todo{Description of the Remote Command Control Center - Adjust it so that it matches the version of CCC realized for VanAssist project}

\begin{figure}
	\centering
	\includegraphics[width=\textwidth, keepaspectratio]{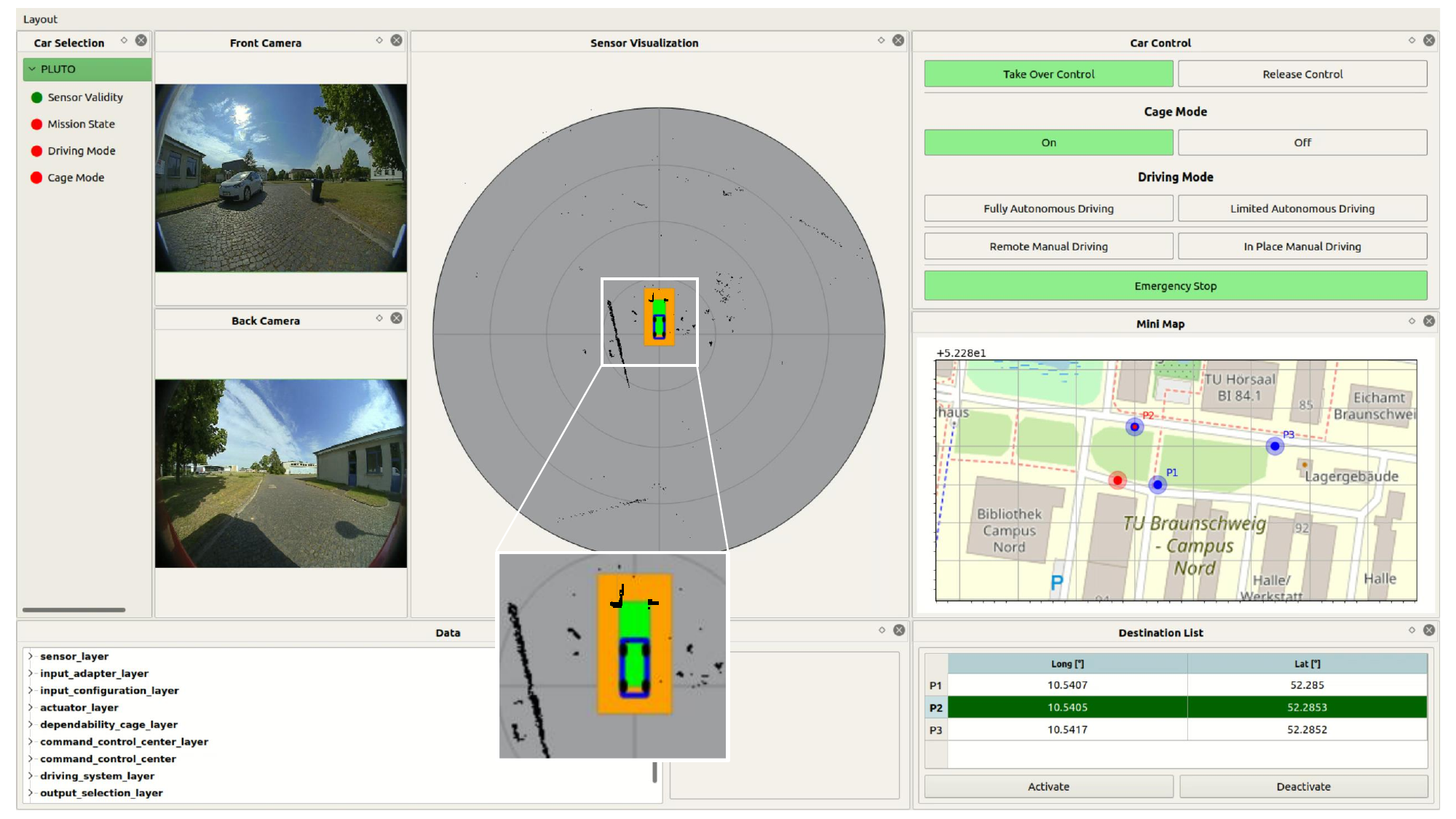}
	\caption{Overview of the Remote CCC GUI.}
	\label{fig:ccc-overview}
\end{figure}

The remote CCC allows the remote safety driver to visualize the state of the autonomous ego-vehicle based on the sensor data received from its LiDAR and camera sensors as well as the inputs received from the connected dependability cage. Figure \ref{fig:ccc-overview} shows an overview of the graphical user interface (GUI) of the remote CCC. 

On the left side of the display in the command control center there is a summary containing the following attributes: (1) the sensor validity, (2) the mission state, (3) the driving mode, and (4) the cage state. The sensor validity is a Boolean flag which represents the assessment made by the qualitative monitor with respect to the validity of the camera input images. Regarding the mission state, a distinction is made between the states inactive, active, blocked, and completed. The \textit{inactive} state means that the ego-vehicle is not currently performing any driving task. The state \textit{active} means that the vehicle is currently carrying out a driving task, which is not yet completed. If a problem occurs during the current driving task ("Fail-Operational Mode"), which prevents the ego-vehicle from completing it, the state \textit{blocked} is inferred. After the vehicle has finished its driving task, the mission state is considered to be \textit{completed}. The driving mode refers to the current driving mode of the ADS, while the cage state indicates whether there are any objects detected inside the vehicle's safe zone or not. All these attributes describe together the state of the ego-vehicle. The possible values of each attribute are listed in Table \ref{tab:ego-vehicle-information}.

\begin{table}[h]
	\centering
	\ra{1.2}
	%\captionsetup[table]{width=13cm}
	\parbox{11cm}{\caption{Information about the Ego-Vehicle displayed in the Remote CCC.}\label{tab:ego-vehicle-information}}
	\begin{tabular}{@{}|p{10em}|p{23em}|@{}}
		\toprule
		\textbf{Attribute Name}&\textbf{Attribute Values} \\
		\midrule
		Sensor Validity & \{Valid, Invalid\} \\
		\midrule
		Mission State & \{Inactive, Active, Blocked, Completed\}\\
		\midrule
		Driving Mode & \{Fully Autonomous Driving, Limited Autonomous Driving, Remote Manual Driving, In-Place Manual Driving, Emergency Stop\} \\
		\midrule
		Cage State & \{Safe Zone Free, Focus Zone Occupied, Clear Zone Occupied\} \\
		\bottomrule		
	\end{tabular}
\end{table}

In the center of the remote CCC display, there is an integrated representation of the LiDAR sensors and the safe zone, which helps the remote safety driver to quickly and intuitively assess the current driving situation of the ego-vehicle. The blue rectangle in the center of the integrated display shows an over-approximated representation of the vehicle's circumference, which is intended to help the safety driver with orientation. Surrounding the representation of the vehicle's circumference is the visualization of the safe zone, which is computed as a function of the vehicle's current speed and steering angle. Therefore, the safe zone increases in size with the vehicle's speed and changes its shape, i.e., rectangle or circle segment, depending on the current steering angle of the ego-vehicle. The green area represents the clear zone and the orange area the focus zone. The black dots surrounding the vehicle represent the point cloud measured by the vehicle's LiDAR sensors. The camera sensor data is visualized to the left of the LiDAR visualization panel. 

Different controls are illustrated on the upper right of the remote CCC display: car controls, cage mode, and driving mode. In the center right corner of the display a mini-map of the ego-vehicle's environment is shown. The list of destinations/missions is displayed on the bottom right remote CCC display.

% ------------------------ END SECTION -------------------------------

% ------------------------ BEGIN SECTION -----------------------------
\section{A Case Study in Parcel Delivery Logistics}
\label{sec:case-study}
The distribution of goods in urban areas is often carried out by large vehicles, i.e., “Sprinter class” vehicles that are used during the last mile of delivery. The classic parcel delivery process involves the postman going door-to-door and stopping often to reach the different customers delivery addresses. Before making the delivery to the end customer, the postman needs to find an appropriate parking spot, which is not always easy in crowded urban areas. After parking his vehicle, the postman removes the parcel from the vehicle and delivers it to the end customer. The postman also needs to bring back on foot any parcels that he could not deliver to the respective end customers. Besides being highly inefficient, the classic parcel delivery process is also prone to cause traffic congestion in urban areas, environmental pollution, as well as wear and tear of the delivery vehicle. 

In order to address the issues mentioned above, the collaborative project VanAssist\footnote{https://www.vanassist.de/} aimed to develop an integrated vehicle and the corresponding system technology that enables largely emission-free and automated delivery of goods in urban centers. The VanAssist project brought together research institutes from four German universities, i.e., Institute for Reliable Embedded Systems and Communication Electronics at HS Offenburg (HSO), Institute for Vehicle Technology (IfF) at TU Braunschweig, Institute for Software and Systems Engineering (ISSE) at TU Clausthal, and Institute for Enterprise Systems (InES) at University of Mannheim, as well as four industrial partners, i.e., BridgingIT GmbH (BIT), DPD Germany GmbH, IAV GmbH, and Ibeo Automotive Systems GmbH. The overall objective of the project was to develop an automated driving system in an electric vehicle, equipped with an intelligent delivery system that is monitored by onboard and off-board monitoring systems. This intelligent delivery vehicle assists the postman, automatically moving to the next delivery point, reducing the postman's effort and enabling continuous movement along the planned route.

This paper presents the contribution of ISSE at TU Clausthal in the VanAssist project. This is the development of a safety concept for automated driving systems, which can handle critical situations or errors and can ensure the safe operation of the automated vehicle. The safety concept consists of two monitoring systems that interact continuously with each other and enable a seamless sharing of responsibility over the dynamic driving task between the automated driving system and the safety driver. These two systems are: (1) an onboard monitoring system (connected dependability cage) that monitors the vehicle and (2) an off-board monitoring system (command control center) that remotely supervises the entire fleet of vehicles as well as the transfer of responsibility over the dynamic driving task between the automated driving system and the safety driver. A detailed presentation of this safety concept is given in Section \ref{sec:solution-concept} of this paper.

% ------------------------ END SECTION -------------------------------

% ------------------------ BEGIN SECTION -----------------------------
\section{Evaluation and Discussion of Results}
\label{sec:evaluation}
%

%----------------------------------------------------------------
% Eval. Introduction

This section discusses the evaluation of the concept of connected dependability cage presented in Section \ref{sec:solution-concept}. In order to evaluate this concept, we defined an overall use case scenario (cf. Section \ref{subsec:evaluation-scenario}). Different sub-scenarios are then extracted from it and used to test the connected dependability cage. We carried out a qualitative evaluation in our lab environment with a model car (cf. Section \ref{subsec:evaluation-lab}) and on a test field track with a full sized car (cf. Section \ref{subsec:evaluation-test-track}). 

%----------------------------------------------------------------
% Eval. Overall Use Case Scenario
%
\subsection{Overall Use Case Scenario}
\label{subsec:evaluation-scenario}
\begin{figure}[htbp!]
	\centering
	\includegraphics[width=0.87\columnwidth]{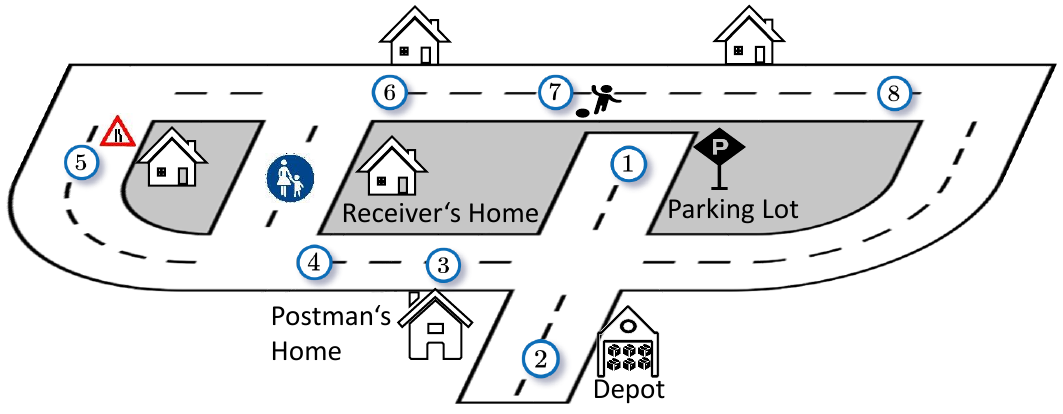}
	\caption{Overview of evaluation scenario}
	\label{fig:scenario_overview}
	% \hspace{0.5em}
\end{figure}

The use case scenario used for the evaluation of our concept is in the application domain of  parcel delivery logistics. A visual overview of the scenario is shown in \Cref{fig:scenario_overview}. 

The scenario consists of several steps. Each step constitutes itself a sub-scenario of the overall use case scenario. In total, the overall use case scenario consists of eight sub-scenarios, which are denoted by unique identifiers 1 to 8. To begin with, the AV drives autonomously from the parking lot (1) to the depot (2), where it picks up packages. From there it drives to the postman's house (3). The postman enters the AV and drives to the home of the first parcel receiver (4). Arriving at the receiver's home, the postman leaves the car for his first delivery round through a pedestrian zone, while the AV drives around the pedestrian zone to meet up with the postman at the first meeting point (6).

On its way to the first meeting point (6), the AV encounters a narrowing in the road and the dependability cage triggers an emergency stop (5). After analyzing the situation, the remote safety driver switches the AV to limited autonomous driving, which limits the speed of the AV and thus uses a smaller safe zone. The AV passes the narrowing using the limited autonomous driving mode and drives to the first meeting point (6). After coming back from his first deliver round, the postman meets with the AV at the first meeting point and retrieves the second batch of parcels out of the AV for his second delivery round. The AV then continues its autonomous drive to the second meeting point (8).

On the way to the second meeting point, children playing ball run on the street and the dependability cage triggers an emergency stop (7). Supervising the situation over the CCC, the remote safety driver waits until the children have left the road, before switching back to the fully autonomous driving mode (7). Once switched to fully autonomous driving mode, the AV continues its trip to the second meeting point (8).

While the AV is waiting for the postman at the second meeting point, another emergency stop is triggered. Analyzing the situation through the sensors visualization panels in the CCC, the remote safety driver recognizes that the front camera is blocked by leaves and informs the postman about this issue (8). Arriving at the second meeting point from its second delivery round, the postman removes the leaves from the camera and gets in the AV (8). The remote safety driver switches the AV back to fully autonomous driving and the AV, together with the postman, drives back to the parking lot (1). This concludes the overview of our overall use case scenario. 

%----------------------------------------------------------------
% Eval. in a Lab Environment
%
\subsection{Evaluation in a Lab Environment}
\label{subsec:evaluation-lab}
\begin{figure}[htbp!]
	\centering
	\includegraphics[width=0.35\columnwidth]{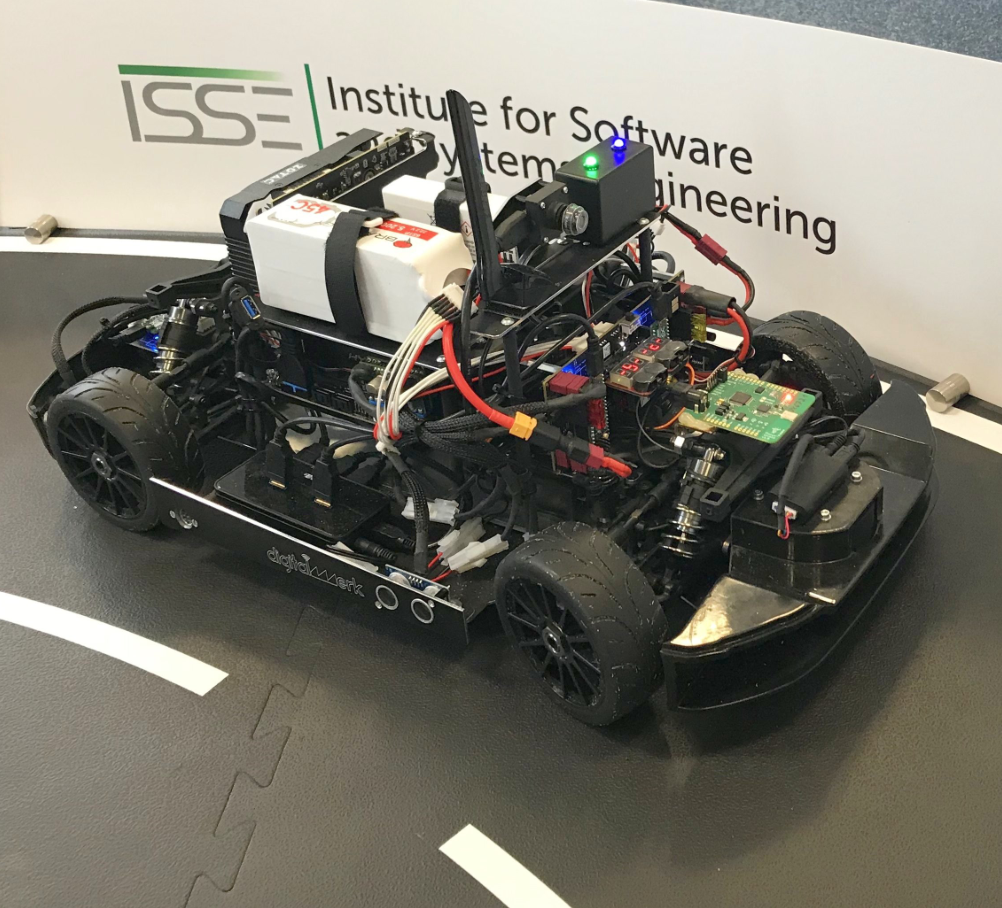}
	\caption{Model car on lab track}% with mats and walls}
\label{fig:model_vehicle}
\end{figure}

For the evaluation in our lab environment, we used a model vehicle with a scale of 1:8 (cf. \cite{Mesow.2017}) equipped with several sensors, which are used to analyze the ego-vehicle state and that of its environment, i.e., LiDAR, camera, ultrasonic sensors, GPS, and IMU. The track was built out of modular black mats of size $1 m \times 1 m$, with street markings and track walls (cf. \Cref{fig:model_vehicle}).

In the rest of this section, we extract three sub-scenarios from the overall use case scenario and show on an exemplary basis how we used these to evaluate the connected dependability cage, i.e., test the different components of the qualitative monitor and the human-machine interaction with the help of the remote CCC.

\subsubsection{Sub-scenario 1: Testing the Human-Machine Interaction.}

\begin{figure}[htbp!]
\centering
\begin{subfigure}{0.45\textwidth}
	\includegraphics[width=\textwidth]{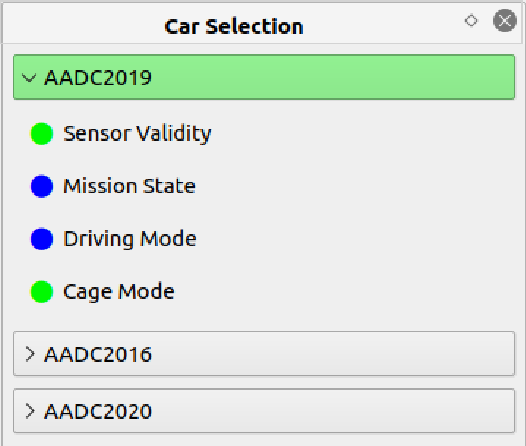}
	\caption{Car Selection panel.}
	\label{fig:ccc_car_selection_s1}
\end{subfigure}
\hfill
\begin{subfigure}{0.45\textwidth}
	\includegraphics[width=\textwidth]{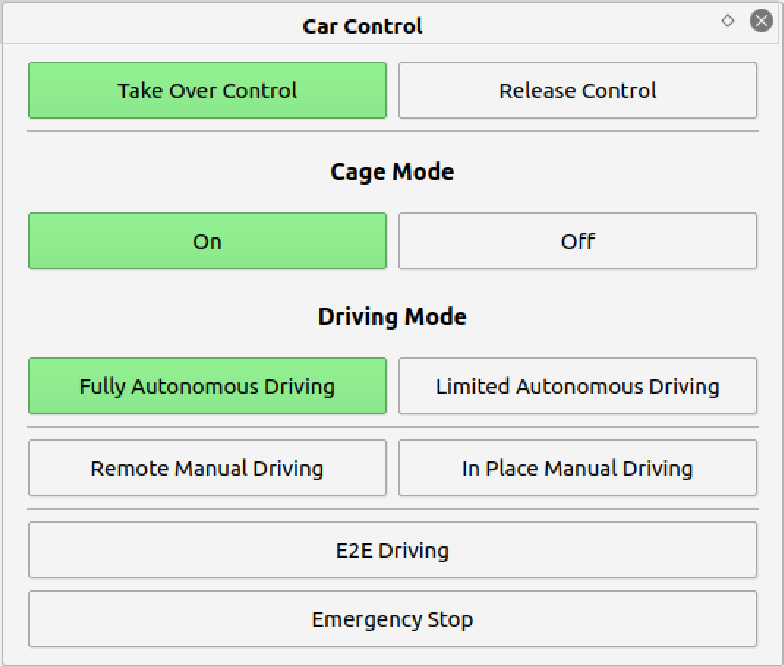}
	\caption{Car Control panel.}
	\label{fig:ccc_car_control_s1}
\end{subfigure}
\caption{
	\label{fig:ccc_car_selection_and_control_s1}%
	Remote CCC - Car Selection and Car Control panels in sub-scenario 1.}
\end{figure}

The remote safety driver uses the different panels of the remote CCC to interact with the AV. In order to start the supervision of an AV, the remote safety driver uses the car selection panel out of the list of AVs displayed on the panel (cf. \Cref{fig:ccc_car_selection_s1}). Once he has selected an AV, the remote safety driver is provided with a very condensed overview of the selected AV's current state through the attributes defined in Table \ref{tab:ego-vehicle-information}. 

A remote CCC can supervise several AVs at a time. However, for the supervision of larger AV fleets it may be necessary to deploy several remote CCCs spanned over a wider area and each having its own jurisdiction. An AV can be controlled by a remote safety driver from a remote CCC only when the control rights over the respective AV are transferred to the remote CCC. The remote safety driver can transfer the control rights to his CCC by using the controls in the car control panel (cf. \Cref{fig:ccc_car_control_s1}). With the control rights over an AV transferred to the remote CCC, the selection of the AV driving modes is also enabled. The remote safety driver has then access to the driving mode selection and and can choose an appropriate driving mode, e.g. fully autonomous driving. This concept enables passing of AV control between different remote CCCs, which are in charge of supervising a large fleet of vehicles.

\begin{figure}[htbp!]
\centering
\begin{subfigure}{0.45\textwidth}
	\includegraphics[width=\textwidth]{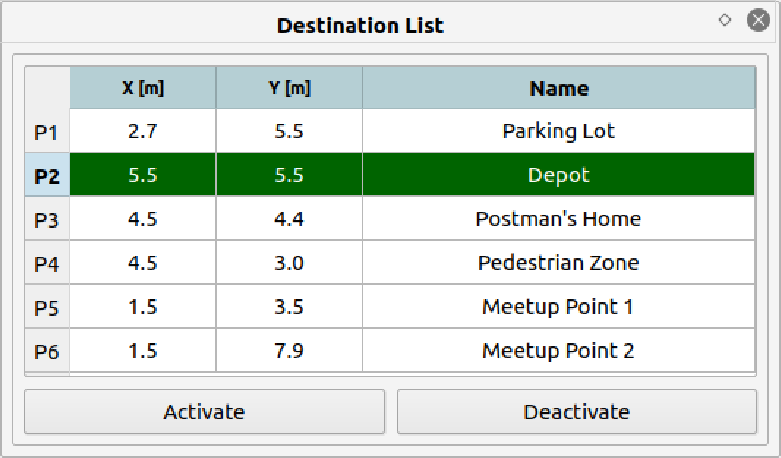}
	\caption{Destination List panel.}
	\label{fig:ccc_destination_list_s1}
\end{subfigure}
\hfill
\begin{subfigure}{0.45\textwidth}
	\includegraphics[width=\textwidth]{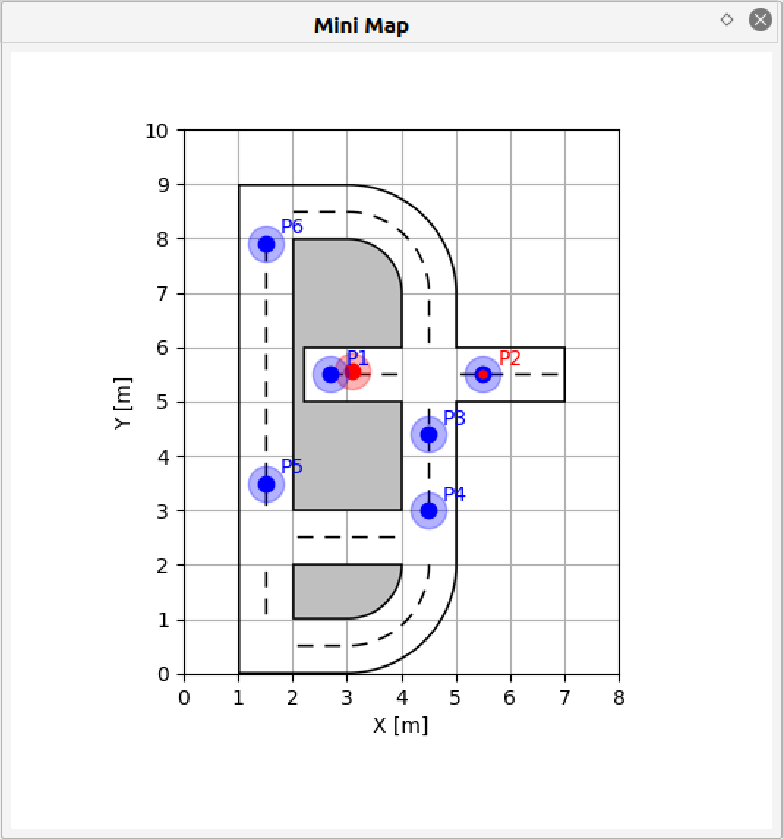}
	\caption{Mini Map panel.}
	\label{fig:ccc_mini_map_s1}
\end{subfigure}
\caption{
	\label{fig:ccc_dest_list_mini_map_s1}%
	Remote CCC - Destination List and Mini Map panels in sub-scenario 1.}
\end{figure}

Before the AV starts driving on the first leg of its trip, the remote safety driver switches the cage on and requests the switch to fully autonomous driving mode (cf. \Cref{fig:ccc_car_control_s1}). The remote safety driver then selects the first destination of the AV out of the destination list panel and activates it (cf. \ref{fig:ccc_destination_list_s1}). In order to track the progress of the AV, the remote safety driver uses the mini map panel to see the current position of the AV (red) and the positions of the destinations (blue) on the track (cf. \Cref{fig:ccc_mini_map_s1}).

\subsubsection{Sub-scenario 7: Testing the Safe Zone and the LiDAR Detector.} 

\begin{figure}[htbp!]
\centering
\begin{subfigure}{0.45\textwidth}
	\includegraphics[width=\textwidth]{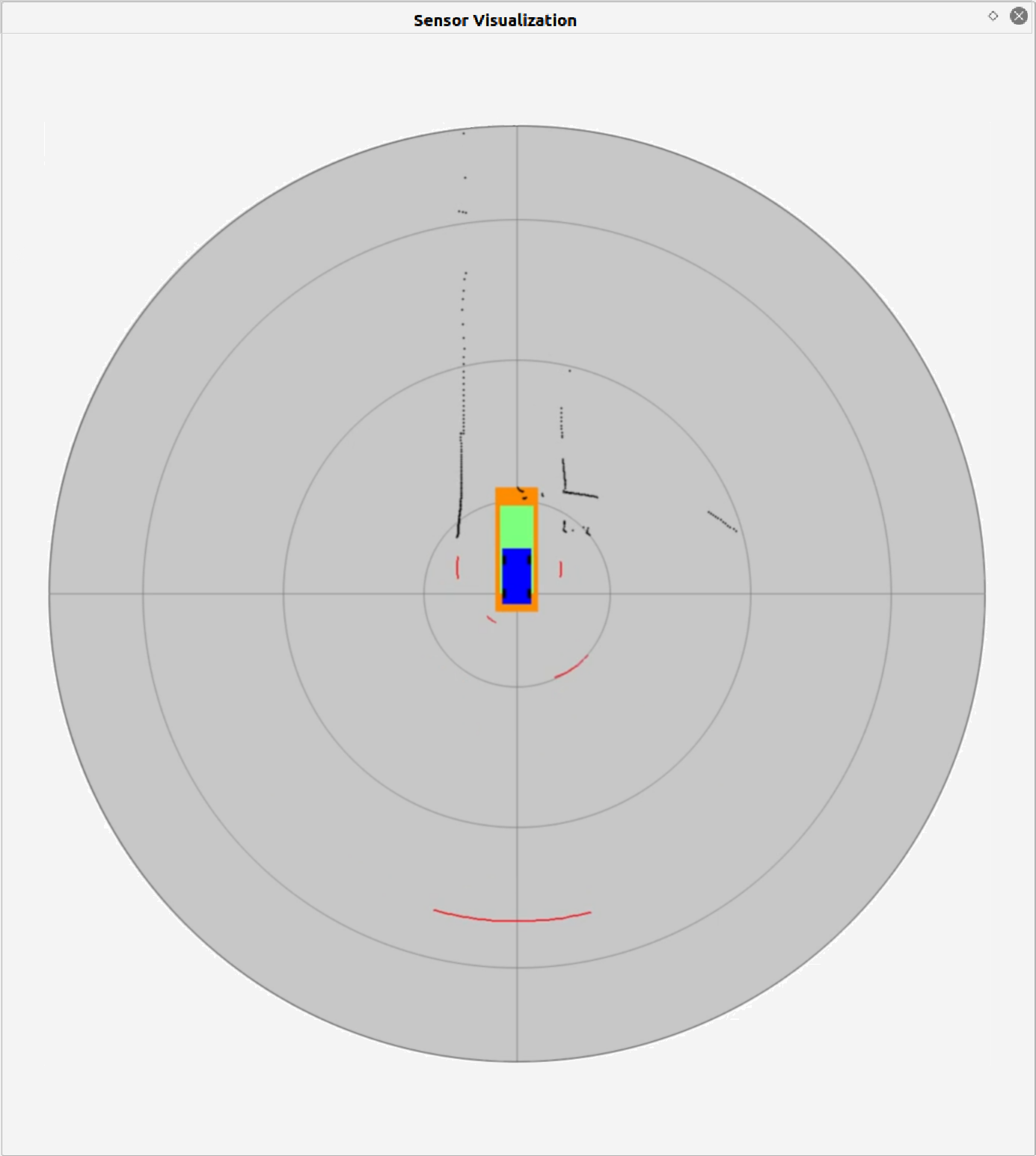}
	\caption{Sensor Visualization panel.}
	\label{fig:ccc_sensor_visualization_s7}
\end{subfigure}
\hfill
\begin{subfigure}{0.45\textwidth}
	\includegraphics[width=\textwidth]{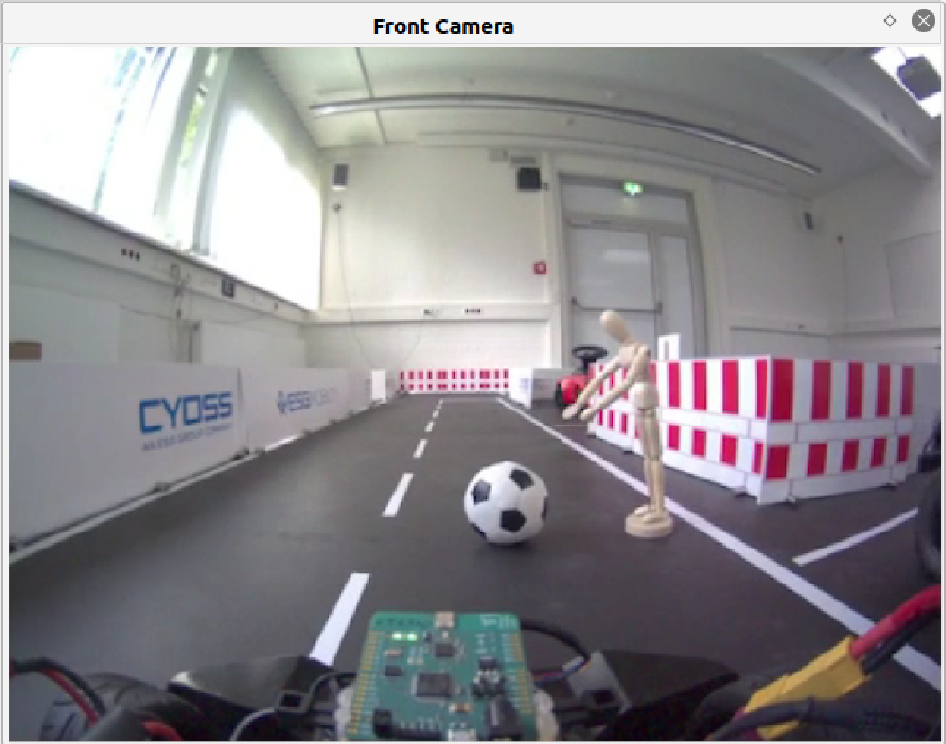}
	\caption{Front Camera panel.}
	\label{fig:ccc_front_camera_s7}
\end{subfigure}

\begin{subfigure}{0.40\textwidth}
	\includegraphics[width=\textwidth]{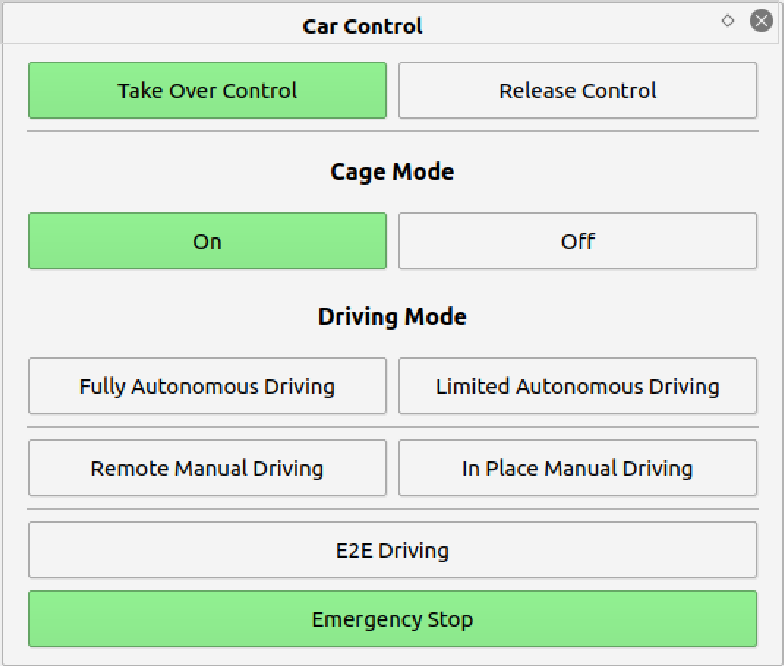}
	\caption{Car Control panel.}
	\label{fig:ccc_car_control_s7}
\end{subfigure}
\caption{
	\label{fig:ccc_sensor_vis_front_camera_car_control_s7}%
	Remote CCC - Sensor Visualization panel, Front Camera panel, and Car Control panel in sub-scenario 7.}
\end{figure}

The safe zone and the LiDAR detector are components of the qualitative monitor, which are used to monitor the safety requirement SR1 by detecting any obstacles in the driving path of the ego-vehicle and trigger an emergency stop to prevent a collision. The remote safety driver is able to visualize the safe zone calculated around the ego-vehicle and the LiDAR points in the sensor visualization panel (cf. \Cref{fig:ccc_sensor_visualization_s7}). In the situation depicted in the sensor visualization panel, there is visible a significant amount LiDAR point inside the safe zone (cf. \Cref{fig:ccc_sensor_visualization_s7}), which leads to the trigger of the emergency stop. Since the AV never switches automatically from emergency stop to fully autonomous driving, it is the responsibility of the remote safety driver to request the switch to fully autonomous driving, once the situation is safe again (cf. \Cref{fig:ccc_car_control_s7}). In addition to the previously described panels, the remote safety driver uses also the camera visualization panel of the CCC to assess the current situation in the AV's environment (cf. \Cref{fig:ccc_front_camera_s7}).

\subsubsection{Sub-scenario 8: Testing the Camera Validator.}

\begin{figure}[htbp!]
\centering
\begin{subfigure}{0.45\textwidth}
	\includegraphics[width=\textwidth]{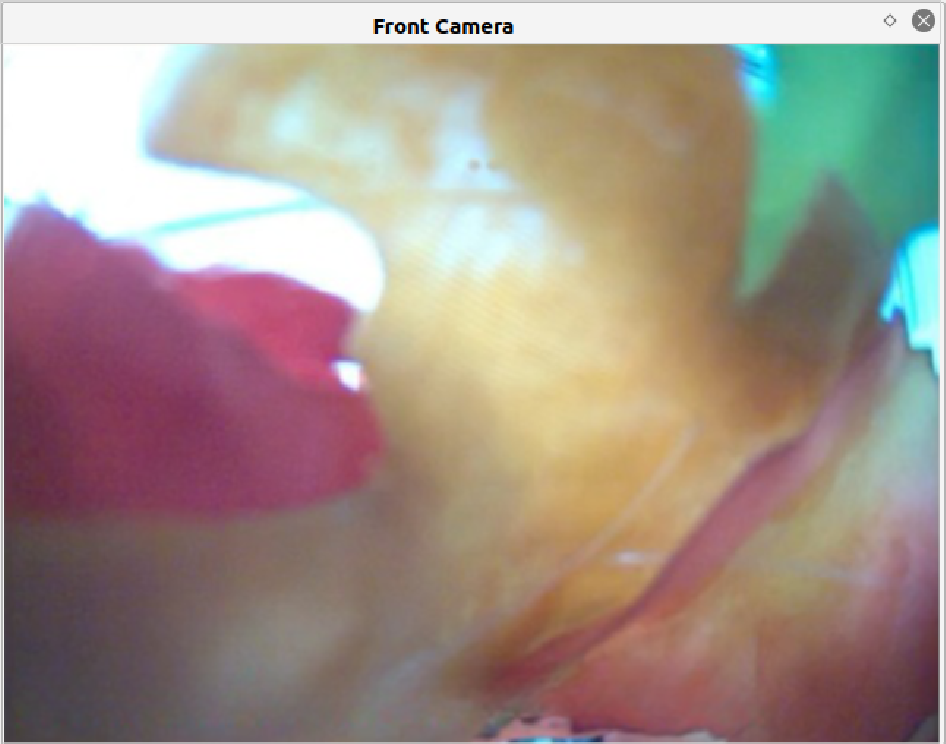}
	\caption{Front Camera panel.}
	\label{fig:ccc_front_camera_s8}
\end{subfigure}
\hfill
\begin{subfigure}{0.40\textwidth}
	\includegraphics[width=\textwidth]{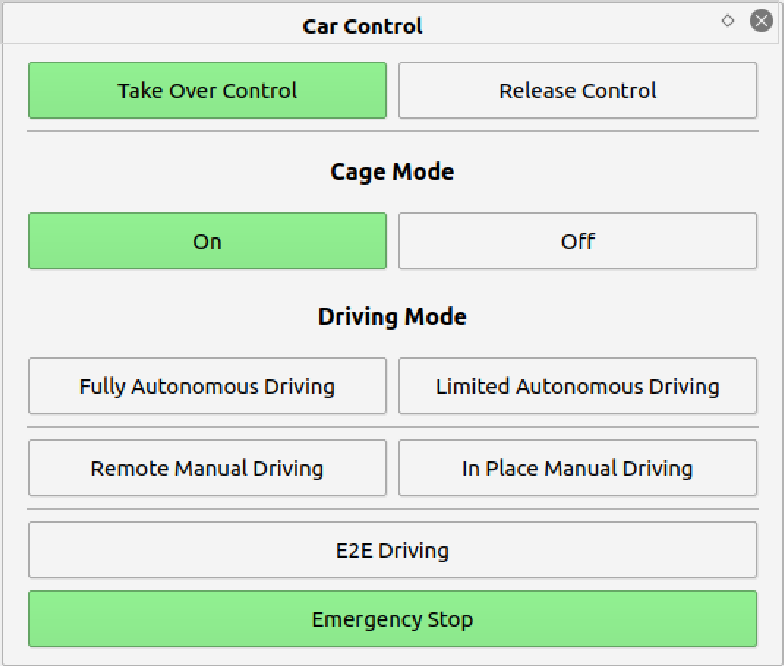}
	\caption{Car Control panel.}
	\label{fig:ccc_car_control_s8}
\end{subfigure}
\caption{
	\label{fig:ccc_front_camera_car_control_s8}%
	Remote CCC - Front Camera panel and Car Control panel in sub-scenario 8.}
\end{figure}

The camera validator is a component of the qualitative monitor, which is used to monitor the safety requirement SR2 by checking the validity of the input camera images. The remote safety driver is able to visualize the status of the camera sensors through the front camera and back camera panels. In the situation depicted in the front camera panel, the front camera is visibly blocked by leaves (cf. \Cref{fig:ccc_sensor_visualization_s7}), which leads to the trigger of the emergency stop (cf. \Cref{fig:ccc_car_control_s8}). Since this situation cannot be resolved remotely, the remote safety driver notifies the postman of this issue and tasks him with solving this issue.

We refer the reader to \cite{Seber.2021} for a more complete description of the lab evaluation that we carried out in the VanAssist project. Additionally, a video which demonstrates and explains the complete lab test scenario can be viewed at \cite{isse.2022}.

%----------------------------------------------------------------
% Eval. on the test field Track
%
\subsection{Evaluation on the Test Field Track}
\label{subsec:evaluation-test-track}
We also evaluated our connected dependability cage concept with a full-size vehicle named PLUTO on a test track located in Braunschweig, Germany.
PLUTO is an electrically-powered full-size vehicle equipped with several sensors, i.e., LiDAR, camera, GPS, and IMU, which  was custom build for the VanAssist project.

The implementation of our connected dependability cage concept for PLUTO presented several new challenges that were especially related to the safe zone component and the LiDAR detector component. One of these challenges was the 360° environment perception around the ego-vehicle provided by eight LiDAR sensors and 4 cameras. The increase in the number of the LiDAR sensors as well as the fact that these were 3D LiDAR sensors led to a significant increase in the volume of LiDAR sensor data and the noise present in these data. Furthermore, PLUTO presented a significantly different vehicle dynamics in comparison to the model vehicle used for the lab evaluation. Last but not least, although the lab track emulated the test field track, the two environments were significantly different from each other due to the fact that the lab track is an indoor environment, while test field track was situated outdoors.

To address these challenges we generalized the safe zone to a circle segment for driving forward and driving backwards. Furthermore, we implemented a z-cutoff to handle the ghost points in the LiDAR data and a clustering algorithm for the detection of objects based on LiDAR data point. We refer the reader to \cite{Helsch.2022} for a more detailed descriptions of these challenges and the implemented solutions.

In the VanAssist project we did not perform a full demonstration of the described overall use case scenario on the test field track, but we were able to carry out test fields for individual sub-scenarios in order to test out the qualitative monitor with its components, i.e., safe zone, LiDAR detector, and camera validator, as well as the human-machine interaction between the remote safety driver and the AV via the remote CCC. 

The tests for the safe zone and the LiDAR detector components were carried out by driving the vehicle PLUTO towards a static obstacle with speeds in the range of ca. \qtyrange{5}{20}{km/h}. We adjusted the calculation of the safe zone as well as the parametrization of the safe zone and the LiDAR detector, so that the emergency stop triggered by the dependability cage brought PLUTO to a full stop at least \qty{1}{m} before the obstacle for the speed range used during the test fields. In addition to this parametrization, we used the noise filtering of the clustering algorithm, in order to filter out LiDAR points that would trigger unnecessary emergency stops of the ego-vehicle. Thus, PLUTO was able to drive multiple rounds around the test track without triggering unnecessary emergency stops. 

The camera validator was tested by placing a piece of cloth in front of the camera and adjusting the parameters of the camera validator for the different cameras and the outdoor lighting conditions.

We used the GUI of the remote CCC described in the previous section in order to carry out the test fields of the human-machine interaction between the remote safety driver and the AV. For a more complete description of the test track evaluation of the VanAssist project we refer to \cite{Seber.2021}.

% ------------------------ END SECTION -------------------------------

% ------------------------ BEGIN SECTION -----------------------------
\section{Summary and Future Work}
\label{sec:summary}
This paper presented an integrated safety concept for safeguarding the safety of ADSs based on the connected dependability cage approach. This approach consists of two runtime monitoring systems: (1) the connected dependability cage which monitors the ADS onboard the ego-vehicle and (2) the remote CCC which is able to supervise off-board an entire fleet of AVs with the cooperation of a remote safety driver. The two runtime monitoring systems are part of an integrated safety architecture for ADSs, which enables the reconfiguration of the ADS and the smooth share and transfer of responsibility over the DDT between the ADS and the remote safety driver. We have carried out a qualitative evaluation of the connected dependability cage approach both in a lab environment using a  model car with 1:10 scale as well as on a test track in Braunschweig using a test vehicle. The results of the qualitative evaluation demonstrated the feasibility of the proposed safety concept for ADSs through its application in scenarios from the domain of parcel delivery logisitcs.

Several directions are interesting in future work. Firstly, the z-cutoff algorithm used in the LiDAR detector component does not always ensure a reliable separation of the LiDAR points pertaining to the ground surface from the rest of the LiDAR data that is relevant for runtime monitoring of the system's ODD. When the dependability cage detects a static obstacle on the road, it triggers immediately an emergency stop, since the safe zone points into the obstacle. This is a fail-safe reaction of the ego-vehicle. In future work, we plan to extend the connected dependability cage so that fail-operational reactions are also possible. Here we envision that the fail-operational reaction could be similar to the reaction of a human driver, who could easily steer back the ego-vehicle, go around the obstacle and continue on its drive. Furthermore, in future we plan to carry out also a quantitative evaluation on a larger set of driving scenarios, which also involve dynamic obstacles in the ego-vehicle environment. In addition, we plan to extend the connected dependability cage approach also with a quantitative monitor, which is able to assess the novelty of the current driving situation of the ego-vehicle.

On an application level, we plan to extend the functionality of the remote CCC so that in addition to the cooperation between the AV and the postman, it also enables the cooperation of the AV with a delivery robot, tasked with receiving the parcels from the AV and delivering them to the end customer. 

% ------------------------ END SECTION -------------------------------

\subsubsection{Acknowledgements}

This research work was made possible through the collaborative project VanAssist, which was funded by the German Federal Ministry of Transportation and Digital Infrastructure (BMVI) under the funding number 16AVF2139E. The project was carried out between October 2018 and June 2021 under the project lead of ZENTEC Center for Technology, Business Start-ups, and Cooperation GmbH. The authors of this paper would like to acknowledge BMVI for the financial support and the valuable collaboration of all project partners involved.
\bibliographystyle{plain} % We choose the "plain" reference style
%\todo{Is "plain" the required bib style? - Yes, it is. See the PDF file authsamp.pdf for this purpose.} 
\bibliography{reference}

% \begin{thebibliography}{6}
% %

% \bibitem {smit:wat}
% Smith, T.F., Waterman, M.S.: Identification of common molecular subsequences.
% J. Mol. Biol. 147, 195?197 (1981). \url{doi:10.1016/0022-2836(81)90087-5                         }

% \bibitem {may:ehr:stein}
% May, P., Ehrlich, H.-C., Steinke, T.: ZIB structure prediction pipeline:
% composing a complex biological workflow through web services.
% In: Nagel, W.E., Walter, W.V., Lehner, W. (eds.) Euro-Par 2006.
% LNCS, vol. 4128, pp. 1148?1158. Springer, Heidelberg (2006).
% \url{doi:10.1007/11823285_121                            }

% \bibitem {fost:kes}
% Foster, I., Kesselman, C.: The Grid: Blueprint for a New Computing Infrastructure.
% Morgan Kaufmann, San Francisco (1999)

% \bibitem {czaj:fitz}
% Czajkowski, K., Fitzgerald, S., Foster, I., Kesselman, C.: Grid information services
% for distributed resource sharing. In: 10th IEEE International Symposium
% on High Performance Distributed Computing, pp. 181?184. IEEE Press, New York (2001).
% \url{doi: 10.1109/HPDC.2001.945188}           

% \bibitem {fo:kes:nic:tue}
% Foster, I., Kesselman, C., Nick, J., Tuecke, S.: The physiology of the grid: an open grid services architecture for distributed systems integration. Technical report, Global Grid
% Forum (2002)

% \bibitem {onlyurl}
% National Center for Biotechnology Information. \url{http://www.ncbi.nlm.nih.gov}

% \end{thebibliography}

\end{document}